\documentclass[letterpaper, 10 pt, conference]{ieeeconf}  

\IEEEoverridecommandlockouts                              

\usepackage{graphicx}
\usepackage{amsmath} 
\usepackage{amssymb}  
\usepackage{booktabs}
\usepackage{multirow}
\usepackage[caption=false]{subfig}
\usepackage[table,xcdraw]{xcolor}
\newcommand{\norm}[1]{\left\lVert #1 \right\rVert}
\usepackage{pifont}
\usepackage[table,xcdraw]{xcolor}
\usepackage[binary-units=true]{siunitx}
\usepackage{footnote}
\usepackage{url}
\usepackage[para,online,flushleft]{threeparttable}

\title{\LARGE \bf
CloudAAE: Learning 6D Object Pose Regression with On-line Data Synthesis on Point Clouds
}

\author{Ge Gao$^{1}$, Mikko Lauri$^{1}$, Xiaolin Hu$^{2}$, Jianwei Zhang$^{1}$ and Simone Frintrop$^{1}$
\thanks{*This work is funded by the German Science Foundation (DFG) in project Crossmodal Learning, TRR 169}
\thanks{$^{1}$Department of Informatics, University of Hamburg, Germany
        {\tt\small \{gao,lauri,zhang\}@informatik.uni-hamburg.de}
        {\tt\small frintrop@informatik.uni-hamburg.de}}%
\thanks{$^{2}$State Key Laboratory of Intelligent Technology and Systems,
Department of Computer Science and Technology,
Tsinghua University, China 
        {\tt\small \{xlhu\}@mails.tsinghua.edu.cn}}%
}

\begin{document}

\maketitle
\thispagestyle{empty}
\pagestyle{empty}

\begin{abstract}

It is often desired to train 6D pose estimation systems on synthetic data because manual annotation is expensive.
However, due to the large domain gap between the synthetic and real images, synthesizing color images is expensive.
In contrast, this domain gap is considerably smaller and easier to fill for depth information.
In this work, we present a system that regresses 6D object pose from depth information represented by point clouds, and a lightweight data synthesis pipeline that creates synthetic point cloud segments for training.
We use an augmented autoencoder (AAE) for learning a latent code that encodes 6D object pose information for pose regression.
The data synthesis pipeline only requires texture-less 3D object models and desired viewpoints, and it is cheap in terms of both time and hardware storage.
Our data synthesis process is up to three orders of magnitude faster than commonly applied approaches that render RGB image data. 
We show the effectiveness of our system on the LineMOD, LineMOD Occlusion, and YCB Video datasets.
The implementation of our system is available at: \url{https://github.com/GeeeG/CloudAAE}.


\end{abstract}

\section{Introduction}
A 6 degrees of freedom (6D) pose is a 3D rotation and 3D transformation between a local object coordinate and a camera or robot coordinate.
Knowing the 6D poses of objects is important for applications such as robotic grasping and manipulation~\cite{zeng2017icra}.
Estimating 6D object pose from images is challenging due to occlusion, background clutter, and different illumination conditions.
According to the recent benchmark challenge for 6D object pose estimation (BOP)~\cite{hodan2020bop}, the performance of deep learning methods has improved compared to the previous year.
Deep learning-based methods benefit from large amounts of high quality training data.
Although there are tools and approaches for annotating the 6D object pose on real images~\cite{marion2018label,xiang2018posecnn}, the annotation for the 6D object pose is expensive.
Moreover, the accuracy of manual labels can not always be guaranteed~\cite{xiang2018posecnn}.

Creating synthetic data eliminates the need for manual labeling and can guarantee the accuracy of labels. 
The difference in appearance between real and synthetic data is called the domain gap. 
The domain gap between real and synthetic RGB images is often large.
In the most recent BOP challenge, the main reason for the performance boost of deep learning-based methods is having additional photo-realistic synthetic images for training~\cite{hodan2020bop}.
These images are created using a physically-based renderer (PBR) and present a much smaller domain gap, compared to naive approaches such as ``render \& paste''~\cite{hodan2020bop}.
Despite being effective, PBR is expensive in the terms of time and hardware storage.
This high cost makes it challenging to scale PBR up to a large number of objects, which is often desired in robotic applications.
Methods using only depth information are more robust in the presence of the domain gap~\cite{hodan2018bop}.
This indicates that if synthesizing data with only depth information, the domain gap is potentially much smaller and easier to fill.
This opens up the possibility to create a lightweight data synthesis pipeline using only depth information.

In this work, we propose a point cloud based 6D pose estimator and a lightweight data synthesis pipeline.
We propose to use an implicit approach during the pose inference stage.
We achieve this by adapting the augmented autoencoder (AAE) proposed by Sundermeyer et al.~\cite{Sundermeyer_2018_ECCV} for a point cloud based system.
By using an AAE, we are able to control which properties the latent code encodes and which properties are ignored~\cite{Sundermeyer_2018_ECCV}.
This is achieved by applying augmentations to the input, and the encoding becomes invariant to those augmentations.
We focus on learning a latent code that encodes the 6D pose information.
The task of the AAE is to reconstruct a point cloud segment in the desired 6D pose.
Moreover, this segment is noise and occlusion free.
In this way, the latent code contains the necessary information for regressing the object pose.
Using the latent code as the input, we use two separate networks for regressing 3D rotation and 3D translation, as suggested in~\cite{wu2017icra,gao2020icra}.
The on-line data synthesis pipeline requires a texture-less 3D object model and the desired viewpoint as the input.
The computational cost is low.

Our contributions are:
\begin{itemize}
    \item We present a new framework for regressing the 6D object pose from point cloud segments. 
    An point cloud based augmented autoencoder is used to learn a latent code that encodes object pose information.
    This code is used for regressing the 6D object pose.
    \item We present a point cloud based lightweight data synthesis pipeline for generating training data.
    Compared to existing RGB based data synthesis systems, the cost of ours is lower in the sense of time and hardware storage.
\end{itemize}{}

We show the effectiveness of the combination of our data synthesis pipeline and pose estimation system on three datasets.
When using only synthetic training data, our model achieves state-of-the-art performance among other synthetic trained methods on the LineMOD dataset~\cite{hinterstoisser2012model}.
Our cheap synthetic point cloud data can replace costly render based synthetic data for training systems using depth for pose inference.

\section{Related Work}
\label{sec:realted_work}
We review deep learning 6D pose estimators on point clouds, unsupervised learning of 6D pose, as well as how training data is synthesized.

\textbf{Deep learning 6D pose estimation on point clouds.}
Frustum PointNets~\cite{qi2018cvpr} uses point cloud segments for pose estimation.
They use PointNet~\cite{qicvpr17} for processing point clouds.
The rotation estimation is formulated as a classification problem, and the result for one angle is reported.
Densefusion~\cite{wangarxiv19} uses PointNet to extract feature vectors from point cloud segments, and a convolutional neural network (CNN) to extract features from the corresponding color information.
Those features are combined for regressing the 6D object pose.
PVN3D~\cite{He_2020_CVPR} adapts the feature extraction pipeline from DenseFusion, and uses a 3D Keypoint detection and Hough voting scheme for 6D pose estimation.
CloudPose \cite{gao2020icra} uses a PointNet like structure to extract features from point clouds, and directly regress to 3D rotation and 3D translation with two separate networks.
Our method is the most similar to CloudPose, as we regress to 6D object poses from point clouds.
However, rather than directly regressing to 6D pose from point clouds, we use the latent code from an AAE as the input to pose regressors.

\textbf{Unsupervised learning of 6D pose.}
The augmented autoencoder is proposed in~\cite{Sundermeyer_2018_ECCV}.
It is a variant of the Denoising Autoencoder~\cite{vincent2010stacked}.
The authors use 2D bounding boxes for translation estimation, and use the AAE for 3D rotation estimation.
For 3D rotation, rather than explicitly mapping from an input image to a pose label, the AAE learns implicit codes of object orientations in a latent space.
A codebook of latent codes is created off-line, and they use a nearest neighbor search to compare a test code within the codebook. 
This approach is improved by~\cite{wen2020ral} by adding edge priors.
Our 6D pose estimation pipeline adapts the AAE concept \cite{Sundermeyer_2018_ECCV} for point clouds.
There are three main differences between our approach and \cite{Sundermeyer_2018_ECCV}.
First, our approach does not require creating off-line codebooks.
Second, we use the latent code from the AAE for both 3D translation and 3D rotation estimation.
Third, they use 2D color image as input while ours uses point clouds.

\textbf{Data synthesis methods for 6D pose estimation.}
Existing data synthesis methods outputs synthetic color images.
To create synthetic data, the viewpoint and the appearance from that viewpoint must be determined.
Object appearance is obtained from either textured 3D object models or real world training datasets.
Viewpoints are selected using some simple or dataset-based heuristics. 
One common approach is to define spheres around the textured 3D object model with fixed radii, and sample viewpoints from a hemisphere that covers the upper part of the object model~\cite{hinterstoisser2012model,zakharov2019dpod}, or from the full sphere~\cite{Sundermeyer_2018_ECCV, wen2020ral}.
The object model is rendered in the target viewpoint onto a plain or random background.
The advantage of this method is that it provides a good coverage for 3D rotation.
The limitation is that using fixed radii puts constraints on the coverage of 3D translation.

The object appearance and viewpoint can also be directly obtained by segmenting foreground objects from an existing training set~\cite{Tekin2018cvpr, rad2017iccv,Park_2019_ICCV}.
More data augmentation, such as background noise, can be applied to the rendered image for reducing the domain gap~\cite{tremblay2018corl:dope}.
However, due to the domain gap, this strategy tends to be insufficient by itself for the 6D object pose estimation task and causes performance drop~\cite{hodan2018bop}.
Moreover, this approach also relies on having the annotated real training data.

Some approaches synthesize data with photo-realistic appearance and physically plausible object poses~\cite{tremblay2018corl:dope,hodan2019photorealistic,Mitash2017iros}.
This approach requires a render to create object appearances with realistic lighting and reflection, and physics simulators to provide physically plausible poses.
Rendering is computationally expensive, and the data created off-line requires large amounts of hardware storage. 
Compared to existing approaches, ours is computationally lightweight, and does not need textured 3D object models or segmented foreground objects from real-world training data.
Another difference is that the existing approaches mostly conduct off-line data generation while ours can be used on-line.

\begin{figure*}[t!]
  \centering
    \includegraphics[width=\textwidth]{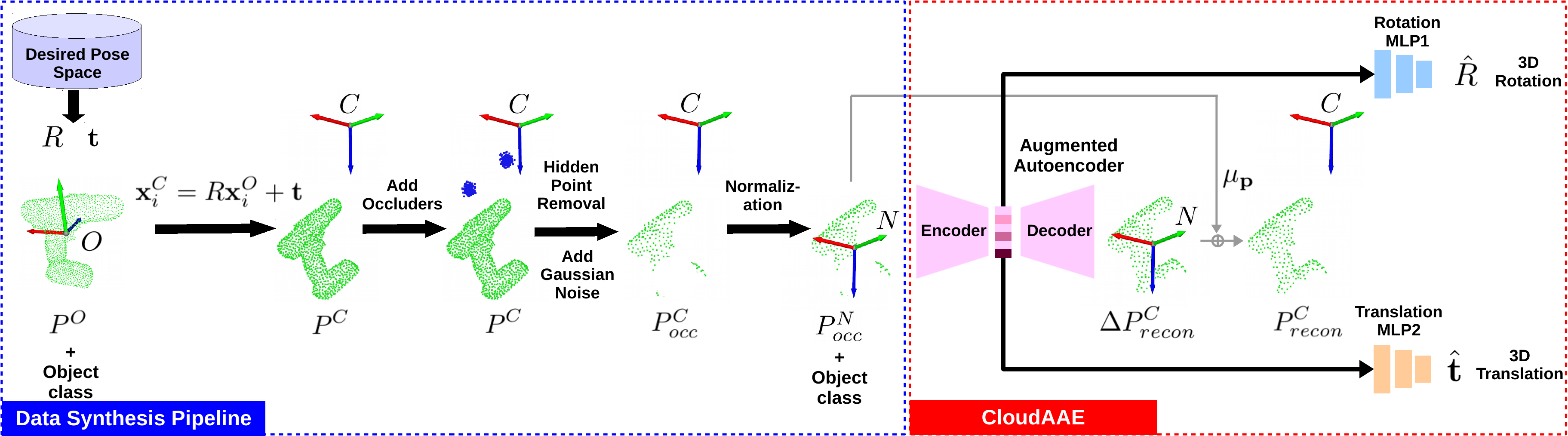}
  \caption{Left, Data synthesis pipeline. The input for our on-line training pipeline is a 3D object model $P^O$ with class information, and the desired 3D rotation $R$ and translation $\mathbf{t}$. $P^O$ is at object coordinate $O$. With $R$ and $\mathbf{t}$, the object model is first transformed to the camera coordinate $C$. The transformed model is denoted by $P^C$. Spherical occluders $S$ (denoted in blue) are added between $C$ and $P^C$. Hidden point removal is applied to $P^C\cup S$ to remove points in $P^C$ and are not visible from $C$.  Zero-mean Gaussian noise is added to each remaining point to introduce variance. The final point set is $P^C_{occ}$. $P^C_{occ}$ is normalized into $P^N_{occ}$, by subtracting the mean $\mathbf{{\mu}_p}$ of $P^C_{occ}$. Right, CloudAAE. The input is $P^N_{occ}$ and the corresponding class information. The expected $\Delta P^C_{recon}$ is a noise and occlusion free segment. By adding $\mathbf{{\mu}_p}$ to $\Delta P^C_{recon}$, we obtain $P^C_{recon}$ at the desired 6D pose. Meanwhile, the latent code is used with two separate 3-layer multi-layer perceptrons (MLP1 and MLP2) for regressing 3D rotation $\hat{R}$ and 3D translation $\hat{\mathbf{t}}$.}
  \label{fig:system_fig}
  \vspace{-2.5mm}
\end{figure*}

\section{Augmented Autoencoder based 6D pose estimation}
\label{sec:system_architecture}
We introduce our data synthesis pipeline and augmented autoencoder (AAE) based 6D pose estimation system, referred to as CloudAAE.
CloudAAE is a multi-class system and we use the same system to predict poses for objects from different classes.
Figure~\ref{fig:system_fig} shows an overview of the proposed system.
Given a known object represented by a set of points in the camera coordinate $C$, the aim of a 6D pose estimation system is to find the translation $\mathbf{t}$ and rotation $R$ that describes the transformation from the object coordinate system $O$ to $C$.
The data synthesis pipeline creates a point cloud segment $P^C_{occ}$, and $P^C_{occ}$ is normalized by removing its coordinate mean $\mathbf{{\mu}_p}$ before further steps.
We denote the normalized $P^C_{occ}$ as $P^N_{occ}$.

By using the data synthesis pipeline, we apply random noise and occlusion to the input point cloud segment to add variance in the training samples.
The AAE reconstructs a noise and occlusion free object segment at the desired 6D pose.
Hence, the latent vector encodes the 6D object pose information, and it is used for 6D pose regression.

\subsection{Data synthesis with point clouds}
\label{subsec:data_synthesis}
The inputs for the data synthesis pipeline are a 3D object model and a 6D pose.
We use 3D object models represented by point clouds.
A 3D rotation $R$ and a 3D translation $\mathbf{t}$ are drawn from a desired pose distribution.
The desired pose distribution can be the poses in a training set~\cite{Tekin2018cvpr, Park_2019_ICCV}.
It can also be a pool of poses covering the same distribution as the poses in a training set~\cite{Su_2015_ICCV, wu2017icra, Manhardt2019cvpr}.
In this work, if the train dataset contains a large amount of poses, we use them as the 6D pose for data synthesis.
Otherwise, we draw poses from the distribution of train poses.

Given an object model represented by the set of points $P^O = \{\mathbf{x}_i^O \in \mathbb{R}^3 \mid i =1,2,\ldots,n \}$, the model is transformed from its object coordinate system $O$ to the camera coordinate system $C$ by
\begin{equation}
    \mathbf{x}_i^C = R\mathbf{x}_i^O + \mathbf{t}.
\end{equation}
$
P^C = \left\lbrace \mathbf{x}_i^C \in \mathbb{R}^3\middle| i=1,2, \ldots, n \right\rbrace 
$
is the transformed model,
where $\mathbf{x}_i^C$ is the $i$th point.
To simulate occlusions, we randomly generate a set of points $S$ as the spherical occluders (denoted in blue in Figure~\ref{fig:system_fig}) between the camera origin $C$ and $P$.
We remove points in $P^C\cup S$ that are not visible from $C$ by applying hidden point removal (HPR)~\cite{katz2007siggraph}.
From the set of visible points, we remove the points belonging to $S$.
To add variance to the training sample, we add zero-mean Gaussian noise with standard deviation $\sigma$ to each remaining point.
In all of our experiments, we use $\sigma=1.3mm$.
The final resulting object segment is $P^C_{occ} = \left\lbrace \mathbf{x}_i \in \mathbb{R}^3 \mid i = 1, \ldots, m \right\rbrace$.



\subsection{CloudAAE: network structure}
\label{subsec:pose_estimator}
CloudAAE contains an augmented autoencoder, and the 6D pose regressors.
We adapted the idea of AAE proposed in ~\cite{Sundermeyer_2018_ECCV} for point clouds, and use an adapted version of the dynamic graph CNN (DGCNN)~\cite{dgcnn} as the encoder.
We additionally add two networks for regressing the 6D poses.

\begin{figure}[t]
\centering
\includegraphics[width = \columnwidth]{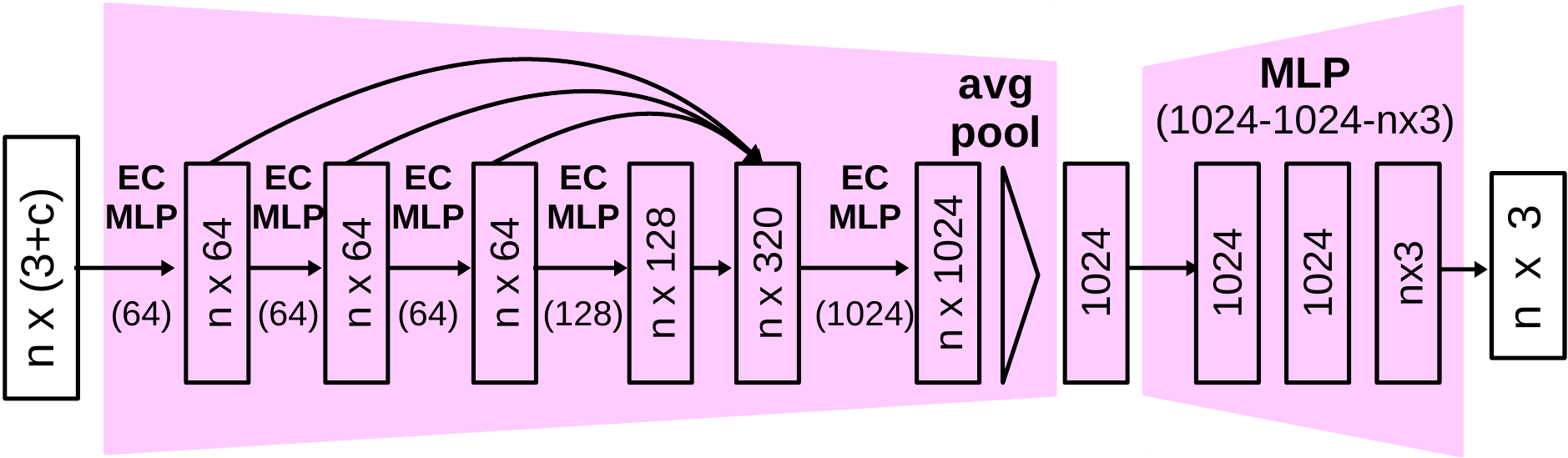}
\caption{Point cloud based Augmented Autoencoder. The numbers of neurons in EC-MLP and MLP layers are indicated in parentheses. EC is short for EdgeConv. Dimensions of intermediate features are indicated without parentheses. Skip connections denote the concatenation of edge features. For the encoder, a 1024-dimensional feature vector for each point with its local neighbors is learned with shared weights. An average pooling layer aggregates the individual features into a 1024-dimensional latent code. Finally, three fully-connected layers output a noise and occlusion free point cloud segment.}
\label{fig:autoencoder}
\vspace{-1mm}
\end{figure}

\textbf{Augmented autoencoder (AAE):}
Figure~\ref{fig:autoencoder} illustrates the architecture of our point cloud based AAE.
The AAE consists of an encoder and a decoder.
The 3D coordinate of each point in $P^N_{occ}$ is concatenated with one-hot class information.
This is used as the input to the encoder.
The encoder computes a latent vector, and the desired output from the decoder is a noise and occlusion free segment in the 6D pose defined by $R$ and $\mathbf{t}$.
The input is of dimension $n\times(3+c)$, in which $3$ represents 3D coordinates, and $c$ is the total number of classes.

Our encoder is an adapted version of DGCNN, which is a deep network processing unordered point sets.
Assume $Q = \left\lbrace \mathbf{q}_i \in \mathbb{R}^p \mid i = 1, \ldots, m \right\rbrace$ is a point set.
$\mathbb{R}^p$ can be a 3D space or an arbitrary feature space.
For each point $\mathbf{q}_i$, a $k$-nearest neighbor graph is calculated.
In all our experiments, we use $k=10$.
The graph contains directed edges $(i,j_{i1}),\dots,(i,j_{ik})$, in which $\mathbf{q}_{j_{i1}},\dots,\mathbf{q}_{j_{ik}}$ are the $k$ closest points to $\mathbf{q}_i$.
For the edge $\mathbf{e}_{ij}$, an edge feature $\begin{bmatrix}\mathbf{q}_i, & (\mathbf{q}_j - \mathbf{q}_i) \end{bmatrix}^T$ is calculated.

The edge features are processed by an edge convolution operation (EdgeConv), which contains an edge function and an aggregation operation~\cite{dgcnn}.
For the edge function, we use an MLP layer with the shared weights for each edge feature.
We use average pooling as the aggregation operation.
This EdgeConv is repeated, calculating the nearest neighbor graph for the feature vectors of the first shared MLP layer, as well as for the subsequent layers.
Finally, the edge features from each EdgeConv are concatenated and processed with an MLP layer.
This concatenation is illustrated with skip connections in Figure~\ref{fig:autoencoder}.
A 1024-dimensional feature vector is learned for each point.
These features are average-pooled to obtain a global representation of the input point cloud segment.
We use four EdgeConv layers on edge features and one EdgeConv layer on the concatenated edge feature.
The encoder outputs a latent code with the dimension of 1024.
The decoder contains 3 fully connected layers with dimensions of 1024, 1024 and $n\times3$.


\textbf{6D pose regressors.}
Since the decoder is able to reconstruct the segment in the same 6D pose from the latent code, the latent code contains the object pose information.
Hence, the latent code is used as the input to two networks for regressing 3D rotation and 3D translation, respectively. 
Each network contains three MLP layers with dimensions 512, 256, and $3$, respectively.


\subsection{CloudAAE: loss function} 
\label{sub:denoising_autoencoder_for_point_cloud}
We normalize $P^C_{occ}$ by removing its mean $\mathbf{{\mu}_p}$ before inputting to the AAE (Figure~\ref{fig:system_fig}).
The output of AAE is the residual 3D coordinates of the reconstructed point cloud segment.
The output of translation prediction is the residual of translation.
The full 3D coordinates and 3D translation are obtained by adding back $\mathbf{{\mu}_p}$.

\begin{figure}[t]
\centering
\subfloat[]{\label{fig:p_vis}\includegraphics[width = 0.43\columnwidth]{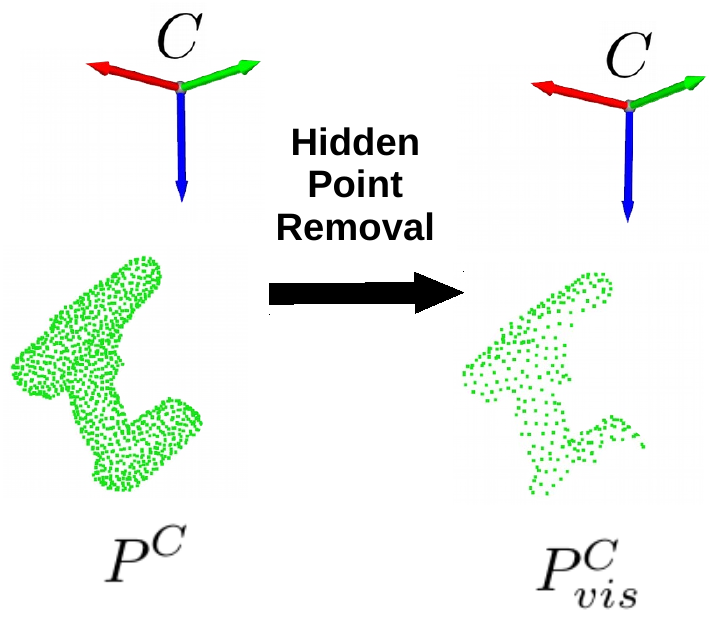}} \hspace{2mm}
\subfloat[]{\label{fig:AE_result}\includegraphics[width = 0.49\columnwidth]{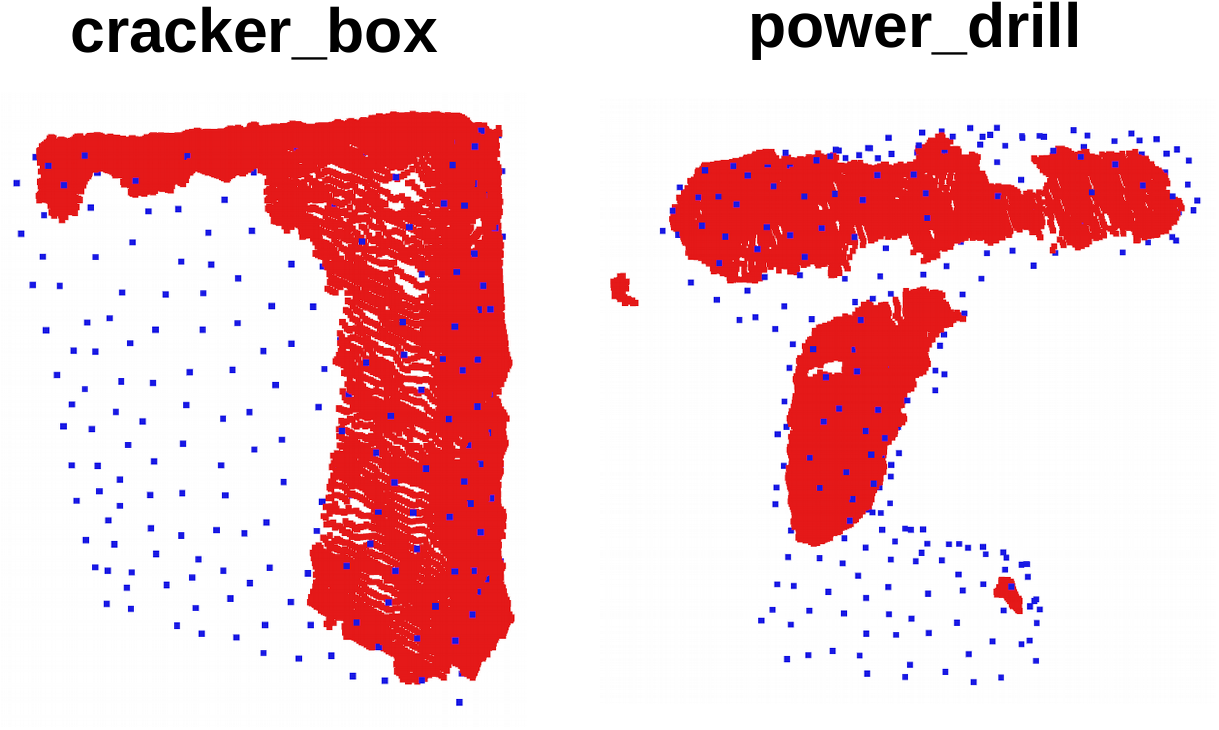}}
\setlength{\belowcaptionskip}{-10pt}
\caption{Illustration of the learning target and the output of AAE. (a) The learning target $P^C_{vis}$. (b) Examples of AAE output for objects in the YCB video dataset during testing. The input to the AAE is denoted in red, and the reconstructed noise and occlusion free point cloud segment is denoted in blue. }
\label{fig:AE_result_fig}
\vspace{-0.5mm}
\end{figure}

\textbf{Augmented autoencoder (AAE).}
Assuming the AAE outputs residual 3D coordinates $\Delta P^C_{recon}$, the full 3D coordinate is
$
    P^C_{recon}= \Delta P^C_{recon} + \mathbf{{\mu}_p}.
$
The learning target $P^C_{vis}$ is obtained by applying HPR to $P^C$ without introducing any occluders, as illustrated in Figure~\ref{fig:AE_result_fig}\subref{fig:p_vis}.
Chamfer distance is used to measure the difference between $P^C_{recon}$ and $P^C_{vis}$,
\begin{align}
l_{CD}(P^C_{recon}, P^C_{vis}) = \frac{1}{m}\sum_{\mathbf{x}\in P^C_{recon}} \min_{\mathbf{y}\in P^C_{vis}} \| \mathbf{x}-\mathbf{y}\|_2 \notag\\
+ \frac{1}{n}\sum_{\mathbf{y}\in P^C_{vis}} \min_{\mathbf{x}\in P^C_{recon}} \| \mathbf{y}-\mathbf{x}\|_2.
\end{align}
The number of points in $P^C_{vis}$ and $P^C_{recon}$ are denoted by $m$ and $n$, respectively.
The desired $P^C_{recon}$ is a denoised, unoccluded object segment. 
Figure \ref{fig:AE_result_fig}\subref{fig:AE_result} shows examples of $P^C_{recon}$ output by a trained AAE.


\textbf{6D pose regressors}.
For rotation regression, the axis-angle representation is used as the regression target.
An axis-angle is a vector $\mathbf{r}\in\mathbb{R}^3$ that represents a rotation of $\theta = \norm{\mathbf{r}}_2$ radians around the unit vector $\frac{\mathbf{r}}{\norm{\mathbf{r}}_2}$.
Geodesic distance is used as the loss function for measuring the distance between prediction $\hat{\mathbf{r}}$ and ground truth $\mathbf{r}$~\cite{gaoeccvw18}.
With $\hat{R}$ and $R$ being the corresponding rotation matrices for $\hat{\mathbf{r}}$ and $\mathbf{r}$ respectively, the \textit{rotation loss function} $l_r(\hat{\mathbf{r}}, \mathbf{r})$ is defined as
\begin{equation}
    l_r(\hat{\mathbf{r}}, \mathbf{r}) = \arccos\left(\frac{\mathrm{trace}(\hat{R}R^T)-1}{2}\right).
\end{equation}

The regression target for translation is the residual of translation.
Given $\widehat{\Delta\mathbf{t}}$ as the translation residual, full translation prediction is $\hat{\mathbf{t}} = \widehat{\Delta\mathbf{t}} + \mathbf{{\mu}_p}$.
With $\mathbf{t}$ being the ground truth translation, the \textit{translation loss function} $l_t(\hat{\mathbf{t}}, \mathbf{t})$ is measured with $L2$-norm
$
    l_t(\hat{\mathbf{t}}, \mathbf{t}) = \norm{\mathbf{t} - \hat{\mathbf{t}}}_2.
$

\textbf{Total loss function}.
The total loss $l_{total}$ used for training the AAE and pose regressors is the combination of the reconstruction and the pose loss:
\begin{equation}
    l_{total} =  \alpha l_{CD} + \beta l_t + l_r,
\end{equation} 
where $\alpha$ and $\beta$ are scaling factors.

\subsection{CloudAAE: testing phase}
In the testing phase, the trained CloudAAE is used for estimating 6D object poses.
If the input is an RGBD image, semantic segmentation is used to obtain the object segment and the corresponding class information.
After segmenting the depth image, the point cloud is obtained using the depth value and the camera intrinsic parameters.
The point cloud segment is normalized and concatenated with one-hot class information.
This is used as the CloudAAE input.
As semantic segmentation is a well-studied topic~\cite{segnet}, we assume the object segment and class information is provided by an off-the-shelf method.
After obtaining the pose estimates, we further refine the poses with Iterative Closest Point (ICP).


\begin{table*}[h!]
\captionsetup{position=top}
\centering
\caption{Percentage of correctly estimated poses on LineMOD and LineMOD Occlusion. E.box and glue are reported with ADD-S, others with ADD. Among the methods trained only on synthetic data, our method has the best results on LineMOD.}
\label{tab:lmandlmo}
\subfloat[LineMOD.]{
\label{tab:lm_add}
\resizebox{0.59\textwidth}{!}{%
\begin{tabular}{@{}ccccccccc@{}}
\toprule
 & \multicolumn{5}{c|}{Synthetic} & \multicolumn{3}{c}{Synthetic+Real} \\ \midrule
\multicolumn{1}{c|}{Modality} & \multicolumn{1}{c|}{RGB} & \multicolumn{1}{c|}{RGBD} & \multicolumn{3}{c|}{D} & \multicolumn{1}{c|}{RGBD} & \multicolumn{2}{c}{D} \\ \midrule
\multicolumn{1}{c|}{Method} & \multicolumn{1}{c|}{\begin{tabular}[c]{@{}c@{}}EEGP-AAE\\ +ICP~\cite{wen2020ral}\end{tabular}} & \multicolumn{1}{c|}{\begin{tabular}[c]{@{}c@{}}SSD-6D\\ +ICP~\cite{kehl2017iccv}\end{tabular}} & \begin{tabular}[c]{@{}c@{}}CloudPose\\ +ICP~\cite{gao2020icra}\end{tabular} & Ours & \multicolumn{1}{c|}{\begin{tabular}[c]{@{}c@{}}Ours\\ +ICP\end{tabular}} & \multicolumn{1}{c|}{PVN3D~\cite{He_2020_CVPR}} & Ours & \begin{tabular}[c]{@{}c@{}}Ours\\ +ICP\end{tabular} \\ \midrule
\multicolumn{1}{c|}{ape} & 87.4 & - & 58.3 & 74.5 & \multicolumn{1}{c|}{\textbf{92.5}} & \textbf{97.3} & 80.2 & 92.5 \\
\multicolumn{1}{c|}{bvise} & \textbf{96.1} & - & 65.6 & 86.6 & \multicolumn{1}{c|}{90.8} & \textbf{99.7} & 85.7 & 91.8 \\
\multicolumn{1}{c|}{cam} & \textbf{91} & - & 43 & 65.6 & \multicolumn{1}{c|}{85.7} & \textbf{99.6} & 61 & 88.9 \\
\multicolumn{1}{c|}{can} & 89.5 & - & 84.7 & 90.2 & \multicolumn{1}{c|}{\textbf{95.1}} & \textbf{99.5} & 93.1 & 96.4 \\
\multicolumn{1}{c|}{cat} & 96.6 & - & 84.6 & 90.7 & \multicolumn{1}{c|}{\textbf{96.8}} & \textbf{99.8} & 94.4 & 97.5 \\
\multicolumn{1}{c|}{driller} & 78 & - & 83.3 & 97.3 & \multicolumn{1}{c|}{\textbf{98.7}} & \textbf{99.3} & 98.2 & 99 \\
\multicolumn{1}{c|}{duck} & 69.4 & - & 43.2 & 50 & \multicolumn{1}{c|}{\textbf{84.4}} & \textbf{98.2} & 62.6 & 92.7 \\
\multicolumn{1}{c|}{e.box} & \textbf{100} & - & 99.5 & 99.7 & \multicolumn{1}{c|}{99.2} & \textbf{99.8} & \textbf{99.8} & \textbf{99.8} \\
\multicolumn{1}{c|}{glue} & 99 & - & 98.8 & 93.5 & \multicolumn{1}{c|}{98.7} & \textbf{100.0} & 94.1 & 99 \\
\multicolumn{1}{c|}{holep} & 66.5 & - & 72.1 & 57.9 & \multicolumn{1}{c|}{\textbf{85.3}} & \textbf{99.9} & 84.4 & 93.7 \\
\multicolumn{1}{c|}{iron} & \textbf{98.8} & - & 70.3 & 85 & \multicolumn{1}{c|}{91.4} & \textbf{99.7} & 89.5 & 95.9 \\
\multicolumn{1}{c|}{lamp} & \textbf{94.4} & - & 93.2 & 82.1 & \multicolumn{1}{c|}{86.5} & \textbf{99.8} & 91.6 & 96.6 \\
\multicolumn{1}{c|}{phone} & 93.5 & - & 81 & 94.4 & \multicolumn{1}{c|}{\textbf{97.4}} & \textbf{99.5} & 93.5 & 97.4 \\ \midrule
avg & 89.2 & 90.9 & 75.2 & 82.1 & \textbf{92.5} & \textbf{99.4} & 86.8 & 95.5 \\ \bottomrule
\end{tabular}
}
}
\hspace{2mm}
\subfloat[LineMOD Occlusion.]{
\label{tab:lmo_adds}
\resizebox{0.25\textwidth}{!}{%
\begin{tabular}{@{}cccc@{}}
\toprule
Train & Mod. & Method & avg \\ \midrule
\multirow{3}{*}{Syn.} & \multirow{3}{*}{D} & \begin{tabular}[c]{@{}c@{}}CloudPose\\ +ICP~\cite{gao2020icra}\end{tabular} & 44.2 \\[1ex]
 &  & Ours & 57.1 \\[1ex]
 &  & \begin{tabular}[c]{@{}c@{}}Ours\\ +ICP\end{tabular} & \textbf{63.2} \\[1ex] \midrule
\multirow{6}{*}{\begin{tabular}[c]{@{}c@{}}Syn.\\ +\\ Real\end{tabular}} & \multirow{2}{*}{RGB} & PoseCNN~\cite{xiang2018posecnn} & 24.9 \\[1ex]
 &  & PVNet~\cite{peng2019pvnet} & 47.25 \\[1ex] \cmidrule(l){2-4} 
 & RGBD & \begin{tabular}[c]{@{}c@{}}PoseCNN\\ +ICP~\cite{xiang2018posecnn}\end{tabular} & \textbf{78.0} \\[1ex] \cmidrule(l){2-4} 
 & \multirow{3}{*}{D} & \begin{tabular}[c]{@{}c@{}}PointVotNet\\ +ICP~\cite{pointvotenet20}\end{tabular} & 52.6 \\[1ex]
 &  & Ours & 58.9 \\[1ex]
 &  & \begin{tabular}[c]{@{}c@{}}Ours\\ +ICP\end{tabular} & 66.1 \\ \bottomrule
\end{tabular}
}
}
\vspace*{-\baselineskip}
\vspace*{-4mm}

\end{table*}

\section{Experiments}
\label{sec:experiments}
We compare our results to the state-of-the-art methods on public datasets.
We also investigate the impact of the amount of synthetic data used for training, as well as different sizes for the latent code in AAE.

\subsection{Datasets and experiment setup}
\label{sub:datasets_and_experiment_setup}
We consider three datasets, the LineMOD dataset (LM)~\cite{hinterstoisser2012model}, LineMOD Occlusion (LMO)~\cite{brachmanneccv14} and YCB video dataset (YCBV)~\cite{xiang2018posecnn}.
Depending on the applicability of the methods, for each dataset, we compare to a subset of state-of-the-art methods SSD-6D~\cite{kehl2017iccv}, EEPG-AAE~\cite{wen2020ral}, CloudPose~\cite{gao2020icra}, PVNet~\cite{peng2019pvnet}, PoseCNN~\cite{xiang2018posecnn}, DenseFusion~\cite{wangarxiv19}, PVN3D~\cite{He_2020_CVPR} and PointVoteNet~\cite{pointvotenet20}.

LM~\cite{hinterstoisser2012model} contains 13 objects from daily scenarios in 13 videos.
Although the target objects are not occluded, this dataset is very challenging for depth based methods due to noise in the depth data~\cite{wen2020ral}.
LMO~\cite{brachmanneccv14} contains 8 objects from the LM dataset.
The objects belong to one video sequence, and have a high level of occlusion.
YCBV~\cite{xiang2018posecnn} contains 21 objects from YCB object set~\cite{calliram15} in 92 videos.
It contains many objects with different degrees of rotational symmetry, as well as occlusions.
It provides both real and synthetic training data.
Compared to the other two datasets, the quality of its depth information is good.

We use the 3D models in point clouds provided by the datasets for the on-line data synthesis.
Tests are conducted with real test images from the official test split.
LM provides approximately $200$ training poses for each class, which is a small amount for generating synthetic data.
To obtain more 6D pose for training, we first calculate a kernel density estimate on the training set poses.
Then we draw $100k$ 6D poses from the distribution for each object class for data synthesis.
For YCBV, we use the $80k$ 6D poses from its synthetic training set as the 6D poses for data synthesis.

For training, we use Adam optimizer with learning rate 0.0008.
The batch size is 128.
The number of points of the input point cloud segments is $n=256$.
Batch normalization is applied to all layers and no dropout is used.
For 6D pose regression, we use $\alpha=1000$ and $\beta=10$, which is given by the ratio between the expected errors of the reconstruction, translation and rotation at the end of the training~\cite{gao2020icra}.
Standard derivation for zero-mean Gaussian noise is $1.3 mm$.
For ICP refinement, we use the simple Point-to-Point registration provided by Open3D~\cite{Zhou2018} and refine for 10 iterations.
The initial search radius is 0.01 meter and it is reduced by $10\%$ after each iteration.
When training with synthetic data, both CloudPose~\cite{gao2020icra} and ours use the proposed data synthesis pipeline.
For testing, our method, CloudPose and DenseFusion~\cite{wangarxiv19} require an off-the-shelf semantic segmentation method.
For LM and LMO, both our method and CloudPose use the test object segmentation provided by the corresponding dataset.
For YCBV, ours, CloudPose and DenseFusion use the object segmentation provided by PoseCNN~\cite{xiang2018posecnn}.
When training with additional real data, we first generate the synthetic segments ($P^O_{occ}$ in Fig.~\ref{fig:system_fig}) off-line and use them with the real data for training.

\subsection{Evaluation Metrics} 
\label{sub:evaluation_metrics}
The average distance (ADD) of model points  and the average distance for a rotationally symmetric object (ADD-S) proposed in~\cite{hinterstoisser2012model} are used as evaluation metrics.
With a set $M$ with $m$ points representing a 3D model, the ADD is defined as:
\begin{align}
\mathrm{ADD}=\frac{1}{m}\displaystyle\sum_{\mathbf{x}\in\mathcal{M}} \norm{ (R\mathbf{x}+\mathbf{t})-(\hat{R}\mathbf{x}+\hat{\mathbf{t}}) }_2.
\end{align}
in which $R$ and $\mathbf{t}$ are the ground truth rotation and translation, and $\hat{R}$ and $\mathbf{\hat{t}}$ are the estimated rotation and translation.
ADD-S is computed using closest point distance.
It is a distance measure that considers possible pose ambiguities caused by object rotational symmetry:
\begin{align}
\mathrm{ADD{\text-}S}=\frac{1}{m}\!\!\displaystyle\sum_{\mathbf{x}_1\in\mathcal{M}} \min_{\mathbf{x}_2\in\mathcal{M}}\norm{ (R\mathbf{x}_1+\mathbf{t})-(\hat{R}\mathbf{x}_2+\hat{\mathbf{t}}) }_2.
\end{align}

For YCBV, a 6D pose estimate is considered to be correct if ADD and ADD-S are smaller than a given threshold $0.1$m.
The area under error threshold-accuracy curve (AUC) for ADD and ADD-S is calculated with maximum threshold $0.1$m.
For LM and LMO Occlusion, a pose is considered correct if the error is less than {10\%} of the maximum diameter of the target object~\cite{hinterstoisser2012model}.


\subsection{Comparison of prediction accuracy} 
\label{sub:comparison}
We first discuss our system performance when using only synthetic data for training on three datasets.
We then briefly discuss the our system performance when using additional real training data.
We divide the result tables into two main parts, separating methods depending on whether real data is used additionally during the training stage.
Furthermore, we also denote the data modality that is used during the pose inference stage.

\paragraph{Synthetic data on LM}
Table~\ref{tab:lmandlmo}\subref{tab:lm_add} shows the evaluation results.
We achieve state-of-the-art performance using only depth for pose inference.
This shows the effectiveness of the combination of our data synthesis pipeline and CloudAAE.
Both ours and CloudPose use depth for pose inference, but our method generalizes better to real test data.
This shows learning a latent code that encodes pose information can improve system robustness.
Using depth for pose inference, our method outperforms RGBD based method SSD-6D.

\paragraph{Synthetic data on LMO}
Table~\ref{tab:lmandlmo}\subref{tab:lmo_adds} shows the comparison results.
Ours outperforms CloudPose by a large margin.
Among the methods without ICP, ours has better performance compared to PVNet and PointVoteNet, which use real data for training.
This shows our system can also handle occlusion.

\paragraph{Synthetic data on YCBV}
Table~\ref{tab:ycbv_adds} shows the results.
Our method is slightly better compared to CloudPose.
Moreover, although trained on synthetic data only, our method has comparable performance to other methods trained on both real and synthetic data.
To the best of our knowledge, we are the first to report results using only synthetic training data on this dataset.


\paragraph{Synthetic and real data}
As shown in Table~\ref{tab:lmandlmo} and~\ref{tab:ycbv_adds}, using additional real training data boosts the performance of our system on all three datasets.
On LM, ours with ICP has close performance to RGBD based PVN3D.
On LMO, among the methods without ICP, ours has the best performance.
With ICP, PoseCNN outperforms our method.
One possible reason is that PoseCNN uses a sophisticated ICP while we use a standard point to point ICP.
Moreover, table~\ref{tab:lmandlmo}\subref{tab:lmo_adds} shows LMO is very challenging for methods that rely on single modality for pose inference, and combining both color and depth is very beneficial.
Another possible factor that is limiting our performance is that, there might be a domain gap between our occlusion simulation and the real data.
On YCBV, we report the results when using real training data with either synthetic data from YCBV (ycbv\_syn) or from our pipeline (our\_syn).
Using real and the two kinds of synthetic data achieves similar performances, and ours is comparable with the RGBD methods~\cite{wangarxiv19,He_2020_CVPR}.

\begin{table}[t]
\centering
\caption{AUC of ADD-S metric on YCB-Video dataset~\cite{xiang2018posecnn}.}
\label{tab:ycbv_adds}
\large
\resizebox{\columnwidth}{!}{%
\begin{tabular}{@{}ccccccc@{}}
\toprule
 & \multicolumn{2}{c|}{Synthetic} & \multicolumn{4}{c}{Synthetic+Real} \\ \midrule
\multicolumn{1}{c|}{Modality} & \multicolumn{2}{c|}{D} & \multicolumn{2}{c|}{RGBD} & \multicolumn{2}{c}{D} \\ \midrule
\multicolumn{1}{c|}{Method} & \begin{tabular}[c]{@{}c@{}}CloudPose\\ +ICP~\cite{gao2020icra}\end{tabular} & \multicolumn{1}{c|}{\begin{tabular}[c]{@{}c@{}}Ours\\ +ICP\end{tabular}} & \begin{tabular}[c]{@{}c@{}}DenseFusion\\ Iterative~\cite{wangarxiv19}\end{tabular} & \multicolumn{1}{c|}{\begin{tabular}[c]{@{}c@{}}PVN3D\\ +ICP~\cite{He_2020_CVPR}\end{tabular}} & \begin{tabular}[c]{@{}c@{}}Ours\\ +ICP\\ (ycbv\_syn)\end{tabular} & \multicolumn{1}{c}{\begin{tabular}[c]{@{}c@{}}Ours\\ +ICP\\ (our\_syn)\end{tabular}} \\ \midrule
avg & 93 & \textbf{93.5} & 93.2 & \textbf{96.1} & 94.0 & 93.6 \\ \bottomrule
\end{tabular}
}
\end{table}


\subsection{Ablation study}
\begin{figure}[t]
\centering
\subfloat[]{\label{fig:amount_vs_accuracy}\includegraphics[width = 0.5\columnwidth]{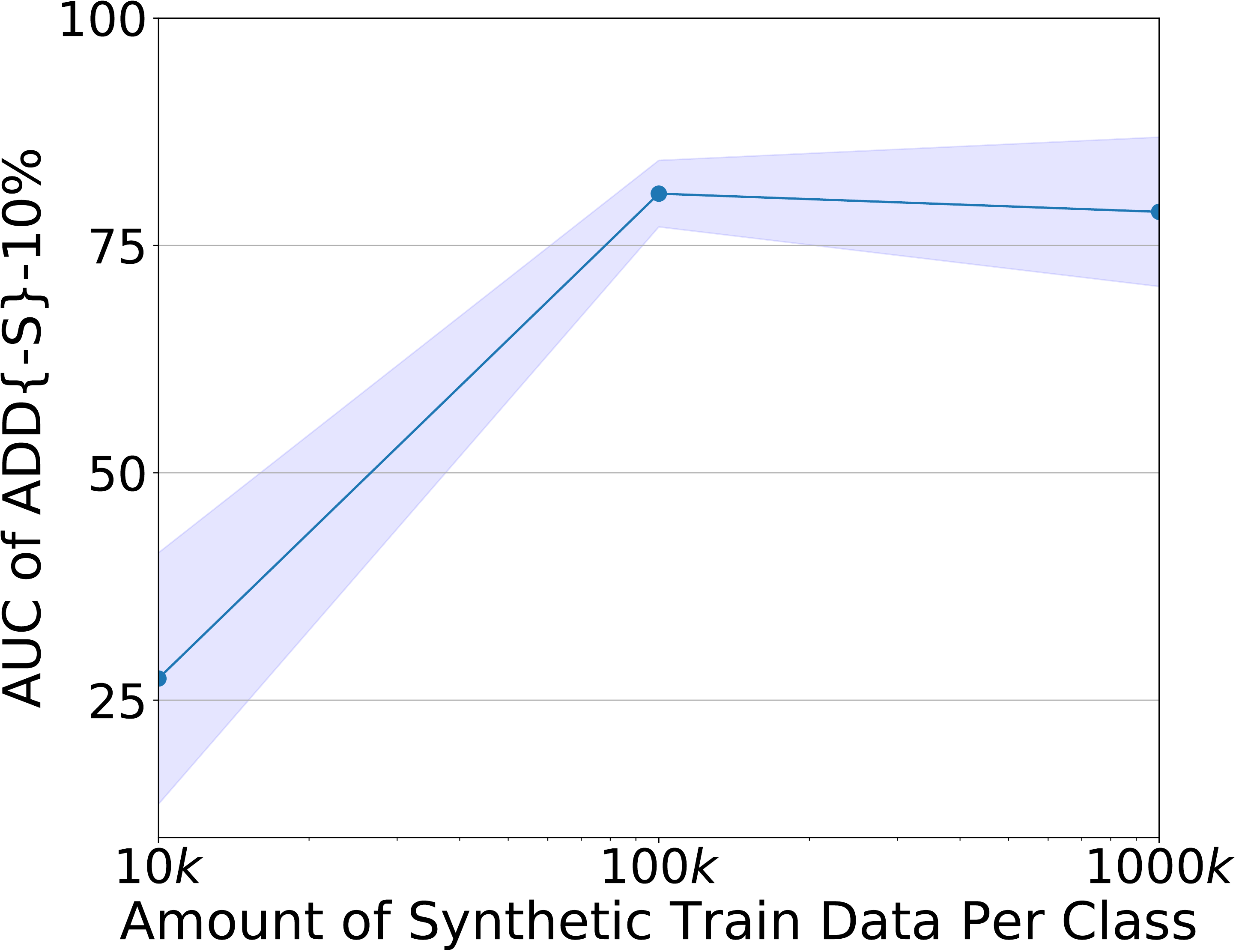}}
\subfloat[]{\label{fig:tsne}\includegraphics[width = 0.45\columnwidth]{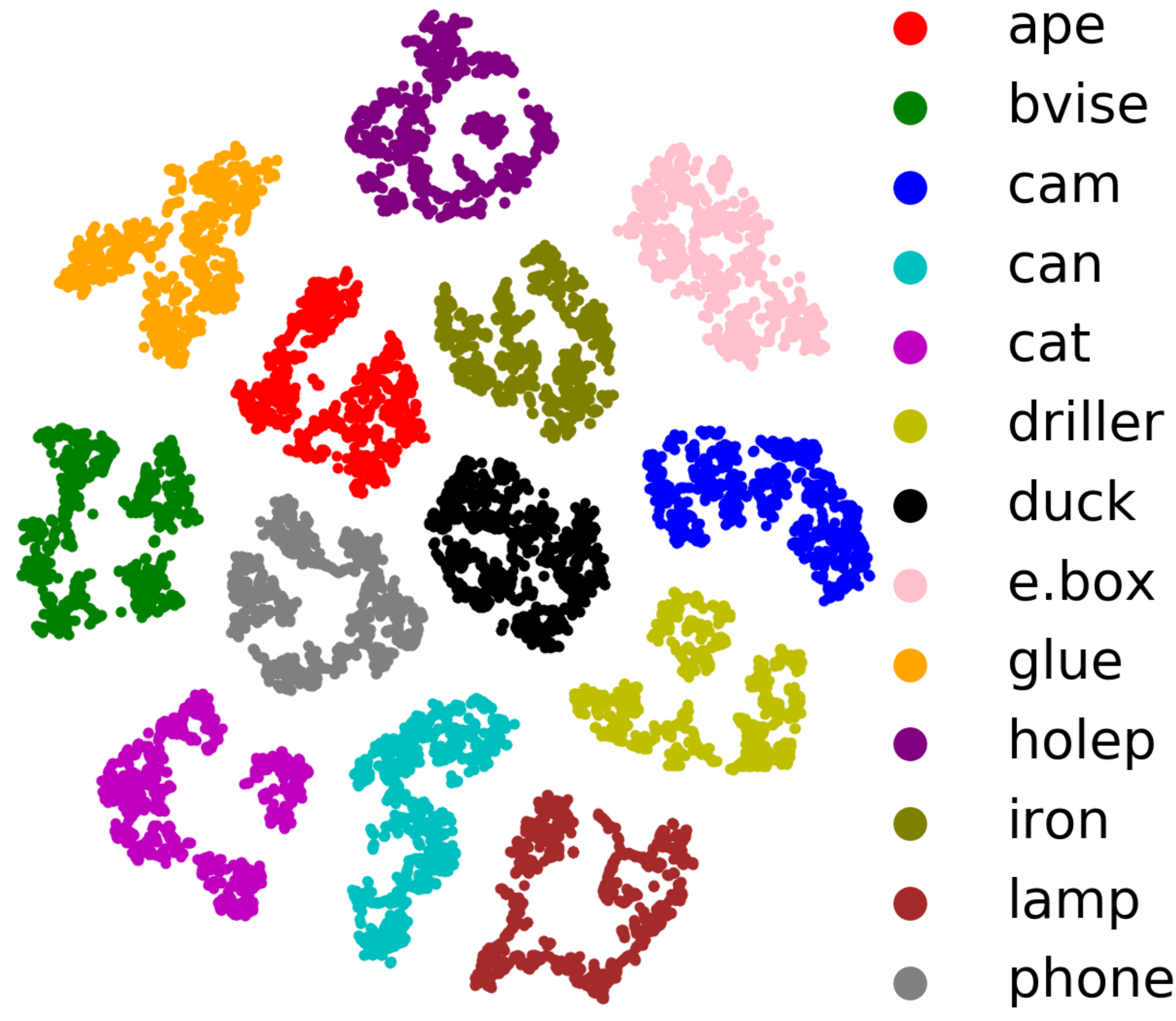}}
\caption{Illustrations of ablation studies. (a) Amount of synthetic training data and performance accuracy (w/o ICP). (b) T-SNE visualization of latent code for the LineMOD dataset.}
\label{fig:ablation}
\end{figure}

We first study how the system performance is impacted by the amount of synthetic training data.
For each object class in LM, we generate either $10k$, $100k$, $1000k$ training poses per class.
For each pose data amount, we train multiple networks with early stopping.
If the training error of the current epoch failed to improve more than $10\%$ of the previous lowest training error, the training process is terminated.
We conducted five trials for $10k$, and three trials for $100k$ and $1000k$.
We generate a set of training data for each trial separately.
The trained networks are evaluated on the test set of LM.
Figure~\ref{fig:ablation}\subref{fig:amount_vs_accuracy} shows the test accuracy (without ICP).
$100k$ samples per class are sufficient for the network to perform well on the test set, and using $1000k$ per class does not provide further performance gain.

We also investigate how the performance is impacted by latent code size.
We train networks with different sizes for the latent code.
All the networks are trained with the same synthetic data and trained until convergence.
The results without ICP are shown in Table~\ref{tab:latent_code_size}.
The generalization ability of a network is the best when the latent code is of size 1024 and 2048.
We pick 1024 for our system.
We show a T-SNE~\cite{JMLR:v9:vandermaaten08a} visualization of the latent code for the LineMOD dataset in Figure~\ref{fig:ablation}\subref{fig:tsne}.
The latent codes are well clustered for each class.

\begin{table}[]
\centering
\caption{Results (w/o ICP) with different latent code sizes on LM.}
\label{tab:latent_code_size}
\begin{tabular}{@{}ccccc@{}}
\toprule
Dimension & 512 & 1024 & 2048 & 4096 \\ \midrule
avg & 82.0 & \textbf{82.1} & \textbf{82.1} & 80.9\\ \bottomrule
\end{tabular}
\vspace*{-\baselineskip}
\end{table}


\subsection{Runtime and hardware storage} 
\label{sub:runtime}
We measure the time performance on a Nvidia Titan X GPU. 
Our system is implemented with Tensorflow.
For the on-line data synthesis, it takes under 30 milliseconds to generate 128 object segments.
The rendering based approach~\cite{denninger2019blenderproc} takes 1-4 seconds to generate an image with 10-20 objects.
Pose estimation by a forward pass through our network takes $0.07$ seconds for a single object. 
The 10 iterations of ICP refinement require an additional $0.02$ seconds.
In the LM experiments, the 3D object models and $13\times100k$ training poses take $\SI{128}{\mega\byte}$ of storage.
If to generate PBR images~\cite{hodan2020bop} with similar amount of objects (e.g. 10 objects per image), $13\times10k$ images would take $\SI{63}{\giga\byte}$.





\section{Conclusion}
\label{sec:conclusion}
We propose a point cloud based lightweight data synthesis pipeline and an augmented autoencoder based system for accurate and fast 6D pose estimation of known objects represented by point clouds.
The data synthesis pipeline is low cost in terms of both time and hardware storage.
Using the synthetic data created by our data synthesis pipeline, our pose estimation system achieves state-of-the-art performance among other synthetic trained methods on a public benchmark.
Moreover, our cheap synthetic point cloud data can replace
expensive render based synthetic data for training systems using depth for pose inference.
Our lightweight data synthesis pipeline enables more agile deployment of object pose estimation systems in robotic applications.





\bibliographystyle{IEEEtran}
\bibliography{root}

\end{document}